\PassOptionsToPackage{dvipsnames,table}{xcolor}
\documentclass{gvc_research}
\usepackage[T1]{fontenc}
\usepackage{lmodern}
\usepackage{multirow}
\usepackage{booktabs}
\usepackage{soul}
\usepackage{pifont}
\usepackage[T1]{fontenc}
\usepackage{graphicx}
\usepackage{amsmath,amssymb}
\usepackage{tikz}
\usetikzlibrary{arrows.meta,positioning,calc,fit,shapes.geometric,decorations.pathreplacing}

\usepackage{makecell}
\usepackage{color, colortbl}
\usepackage{caption}
\usepackage{algorithm}

\usepackage{xspace}
\makeatletter
\DeclareRobustCommand\onedot{\futurelet\@let@token\@onedot}
\def\@onedot{\ifx\@let@token.\else.\null\fi\xspace}

\def\ie{\emph{i.e}\onedot}

\makeatother
\usepackage{array,tabularx}
\usepackage[shortlabels,inline]{enumitem}
\definecolor{tableheaderblue}{HTML}{E8F0FA}
\definecolor{tablegroupgray}{HTML}{F2F2F2}
\definecolor{tabletextgray}{HTML}{555555}
\providecommand{\abstractlist}{}

\usepackage{newtxtext}

\usepackage{algpseudocode}
\definecolor{catgray}{gray}{0.92}

\definecolor{FutureOrange}{HTML}{ff4a05}
\definecolor{teal}{HTML}{e37133}
\definecolor{orange}{HTML}{ff4a05}
\definecolor{purple}{HTML}{e37133}
\definecolor{blue}{HTML}{0d00b2}

\title{Coding4Games: Can Modern Coding Agents Play Modern Video Games?}

\author[1,2]{Shifang Zhao}
\author[3]{Haoyu Wang}
\author[3]{Yuyang Yin}
\author[4]{Dong Chen}
\author[1,2]{Xingyilang Yin}
\author[2]{Yunkang Tao}
\author[1]{Honghao Chen}
\author[1]{Guangzhi Wang}
\author[1]{Yiran Wang}
\author[1]{Junfu Pu}
\author[1]{Cheng Cheng}
\author[1,2,\dagger]{Xiaodong Cun}
\author[1]{Ying Shan}
\affiliation[1]{\textbf{Tencent ARC Lab}}
\affiliation[2]{\textbf{GVC Lab, Great Bay University}}
\affiliationbreak
\affiliation[3]{\textbf{Beijing Jiaotong University}}
\affiliation[4]{\textbf{Nanjing University}}
\contribution[\dagger]{Corresponding author}
\adobedata[Project Page]{\href{https://coding4game.github.io}{\nolinkurl{https://coding4game.github.io}}}

\begin{document}
\maketitle
\enlargethispage{1.65\baselineskip}
\vspace{-4pt}
\begingroup
  \centering
  \captionsetup{hypcap=false,skip=3pt}
  \includegraphics[width=\linewidth]{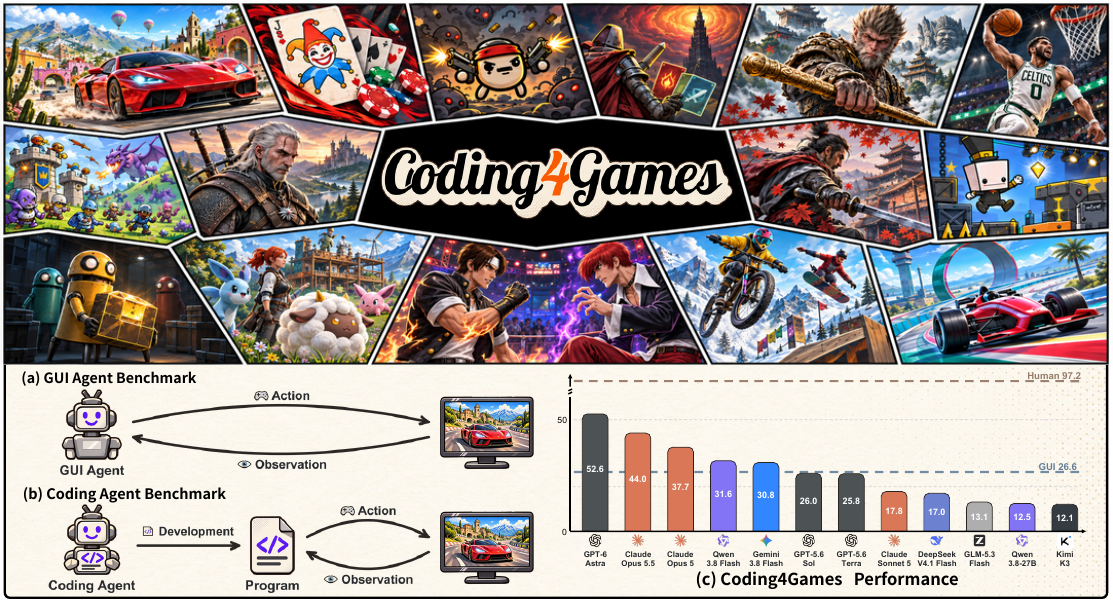}
  \captionof{figure}{\textbf{From playing games directly to building standalone game-playing programs.} (a) In prior computer-use benchmarks, agents play directly through the GUI, querying a language model to determine each action. (b) \mbox{Coding4Games} introduces a different paradigm: a coding agent interacts with the standard player interface to develop a game-playing program. Once finished, the program is frozen and plays autonomously without language-model inference. (c) Average normalized performance of twelve coding agents, compared with human and GUI-agent baselines across 15 games.}
  \label{fig:teaser}
\endgroup
\vspace{4pt}

\abstract{Games, especially modern AAA games, challenge language models to perceive complex scenes, control actions, and make decisions over long horizons. Existing game benchmarks primarily test language models as computer-use agents that query a language model throughout play, incurring repeated inference cost and, in real-time games, acting while the environment continues to evolve. 
We ask whether coding agents can learn from game interaction to develop standalone game-playing programs that run without language-model inference.
We thus introduce \textit{\mbox{Coding4Games}}, a benchmark for evaluating whether coding agents can develop standalone programs that play modern games through visual observations and standard game inputs. Coding4Games has three defining features: \textit{1) Diverse Modern Games}, covering 15 games with diverse visual, strategic, and control demands;
\textit{2) Development--Test Separation}, in which a coding agent uses a sandbox and the standard player interface to iteratively develop a program, after which the selected program is frozen and independently tested without language-model inference; and \textit{3) Outcome and Process Evaluation}, which measures not only game performance and reliability but also how agents use feedback, revise programs, and resolve problems during development. Across 15 games, the best program per game outscores the strongest GUI baseline on 13 and ties it on one. The human player outscores all agents on 10 games. We additionally examine the effects of development budgets. \mbox{Coding4Games} evaluates whether coding agents can turn their understanding of complex interactive environments into programs that work reliably on their own.
}
\section{Introduction}
\label{sec:intro}

Coding agents can autonomously tackle a wide range of software tasks~\citep{jimenez2024swebench,yang2024sweagent,merrill2026terminalbench,openai2025codexcli,anthropic2025claudecode}.
Yet existing evaluations primarily focus on conventional software tasks, leaving their ability to develop programs for complex, dynamic environments less explored. Modern video games present such challenges: their rich visual interfaces, large state and action spaces, and real-time dynamics require programs that integrate perception, strategic decision-making, and precise control.
Developing such programs requires agents to understand game
rules and iteratively debug behavior through interaction,
making modern games a demanding testbed for evaluating
coding agents.


However, existing game benchmarks largely evaluate direct gameplay rather than the development of standalone game-playing programs.
They typically adopt a computer-use setting,\ie, a GUI Agent, in which a language model repeatedly observes the game and selects the next action~(Fig.~\ref{fig:teaser}(a))~\citep{paglieri2025balrog,zhang2026videogamebench,park2026orak,ouyang2026gameworld,zhang2026gameverse}. 
This requires repeated model inference during play;
in real-time games, the environment may continue to evolve during model inferences~\citep{zhang2026videogamebench,ouyang2026gameworld}. A complementary line of work studies API-based game-playing programs, which receive structured game states and issue actions through programmatic interfaces. CodeClash~\citep{yang2025codeclash} and ProxyWar~\citep{peng2026proxywar} ask coding agents to write and iteratively revise bots that compete in game arenas, and related work optimizes game policies or world models represented as code~\citep{kuang2025generativecode,bachrach2025llmpsro,lehrach2026codeworldmodels}. 
However, game-specific APIs expose structured state and action abstractions that bypass the perception and low-level control required in real games, limiting applicability to and transfer across modern titles~(Tab.~\ref{tab:benchmark-comparison}).

\suppressfloats[t]
\begin{table*}
\centering
\footnotesize
\providecolor{checkgreen}{HTML}{237544}
\providecolor{crossred}{HTML}{AB4444}
\providecolor{tablerankwarm}{HTML}{DC9559}

\newcommand{\yes}{\textcolor{checkgreen}{\ding{51}}}
\newcommand{\no}{\textcolor{crossred}{\ding{55}}}
\newcommand{\hd}[1]{{\bfseries #1}}
\setlength{\tabcolsep}{2.5pt}
\renewcommand{\arraystretch}{1.1}
\caption{\textbf{Comparison of game benchmarks.} Coding4Games evaluates coding agents that develop standalone programs for modern video games, with low-level GUI inputs and stepped execution throughout development and evaluation. Target: evaluation target; Setting: game setting; $N$: number of games; Exec.: gameplay executor; Vis.: visual input; Actions: action interface; Mode: execution mode; Reset: episode reset; Test: separate test.}
\vspace{-1em}
\label{tab:benchmark-comparison}
\begin{tabularx}{\linewidth}{@{}
  >{\hsize=2.1\hsize\raggedright\arraybackslash}X
  >{\hsize=1.5\hsize\raggedright\arraybackslash}X
  >{\hsize=.9\hsize\centering\arraybackslash}X
  >{\hsize=.35\hsize\centering\arraybackslash}X
  >{\hsize=.9\hsize\centering\arraybackslash}X
  >{\hsize=.45\hsize\centering\arraybackslash}X
  >{\hsize=1.2\hsize\centering\arraybackslash}X
  >{\hsize=1.0\hsize\centering\arraybackslash}X
  >{\hsize=.8\hsize\centering\arraybackslash}X
  >{\hsize=.8\hsize\centering\arraybackslash}X
@{}}
\toprule[0.8pt]
\rowcolor{tableheaderblue}
\textbf{Benchmark} & \hd{Target} & \hd{Setting} & \hd{$N$} & \hd{Exec.} & \hd{Vis.} & \hd{Actions} & \hd{Mode} & \hd{Reset} & \hd{Test}\\
\midrule
BALROG~\citep{paglieri2025balrog} & Online play & Text / grid & 6 & LM & \yes & Semantic & Stepped & \yes & \no\\
VideoGameBench~\citep{zhang2026videogamebench} & Online play & Retro & 10 & LM & \yes & GUI & Both & NR & \no\\
Orak~\citep{park2026orak} & Online play & Mixed & 12 & LM & \yes & Game API & NR & \yes & \no\\
GameVerse~\citep{zhang2026gameverse} & Video reflection & Mixed & 15 & LM & \yes & GUI / Sem. & Subset & NR & \no\\
GameWorld~\citep{ouyang2026gameworld} & Online play & Web & 34 & LM & \yes & GUI / Sem. & Both & \yes & \no\\
\midrule
CodeClash~\citep{yang2025codeclash} & Code evolution & Arenas & 6 & Program & \no & Program API & N/A & \yes & \no\\
ProxyWar~\citep{peng2026proxywar} & Code generation & Arenas & 9 & Program & \no & Program API & N/A & \yes & \yes\\
\midrule
\rowcolor{tablerankwarm!30!white}
\textbf{Coding4Games (ours)} & \textbf{Coding agents} & \textbf{Modern} & \textbf{15} & \textbf{Program} & \yes & \textbf{GUI} & \textbf{Live} & \yes & \yes\\
\bottomrule[0.8pt]
\end{tabularx}
\vspace{-1em}
\end{table*}

We bring these two lines of work together and ask: \textit{Can a general coding agent understand a modern game through visual interaction and write a program that plays it using only standard game controls, without access to game-specific APIs?}
We define a game-playing program as an executable program built with general-purpose libraries, whose logic maps visual observations to control actions without language-model inference during play, as shown in Fig.~\ref{fig:teaser}~(b). 
Once developed, such a program runs without further language-model calls, while its behavior can be inspected, rerun, and systematically tested. 
This setting evaluates whether coding agents can translate
game understanding into reliable executable behavior.

To answer this question, we introduce \textit{\mbox{Coding4Games}}, a benchmark with three defining features. \textit{1) Diverse Modern Games.} Coding4Games evaluates 15 games spanning diverse visual, strategic, and control demands, rather than restricting evaluation to classic games with comparatively simple visual interfaces and control schemes. It covers games across five categories: Combat \& Action Timing, 3D Sports \& Precision Control, Movement \& Timing, Strategy \& Resource Planning, and Multi-step Task Execution. 
\textit{2) Development--Test Separation.}
During the \emph{Development stage}, coding agents explore games, develop programs, and self-test in a persistent sandbox, interacting with games through a gym-like interface provided by the game server.
During the \emph{Test stage}, the selected program is frozen and evaluated in a fresh environment without language-model inference, testing whether it can play reliably on its own.
\textit{3) Outcome and Process Evaluation.}
Beyond game performance, 
we assess how coding agents use
execution feedback, revise programs, and test whether revisions resolve targeted problems.
We further examine the correlation between the scores achieved in the coding agent and GUI agent settings using the same model.

Our experiments with 12 coding agents demonstrate the potential to turn game interaction into standalone programs. The best program per game outscores the strongest GUI baseline on 13 of 15 games and ties it on one. The human player outscores all agents in 10 games, while the best program per game beats the human in three and ties in two. Process analysis further reveals that self-testing often fails to translate feedback into effective strategy revisions. Together, these findings highlight the promise of code as a reusable representation of interactive behavior, while identifying strategy discovery and reliable improvement through interaction as central challenges for coding agents.

In summary, our contributions are as follows:
\begin{itemize}
    \item We formulate game playing by coding agents as the development of standalone programs through the standard player interface, with a protocol that separates development from independent testing without language-model inference.
    \item We build \mbox{Coding4Games}, an extensible, gym-like environment that brings 15 modern games with diverse perception, control, and planning demands under a standard interface, with standardized task initialization and game-specific scoring.
    \item We evaluate 12 coding agents through game performance and process measures of program grounding, modification depth, and problem resolution, revealing a persistent gap between identifying failures through self-testing and translating feedback into effective strategy.
\end{itemize}

\section{Related work}

\noindent\textbf{Game Agents.}
Deep reinforcement learning and self-play have enabled strong game-playing policies in Atari, board games, and real-time strategy \citep{mnih2013atari,silver2017alphazero,vinyals2019alphastar}, while learning from gameplay demonstrations supports instruction-conditioned and generalist policies \citep{baker2022vpt,simateam2024scaling,magne2026nitrogen}. Language-model agents interleave reasoning and action, revise plans after execution failures, and retain experience in memory \citep{yao2023react,wang2023deps,wang2023jarvis1}. Systems such as Cradle and Voyager further organize behavior into executable skills that agents can compose and reuse \citep{tan2025cradle,wang2023voyager}.
The approaches closest to our setting use language models to revise Atari policies from execution traces, generate best-response programs against Checkers bots, or synthesize executable simulators for classical planning~\citep{kuang2025generativecode,bachrach2025llmpsro,lehrach2026codeworldmodels}.
These approaches separate program construction from execution, allowing the resulting programs to run without per-action language-model inference.
Building on this separation, Coding4Games independently tests the standalone game-playing programs produced by general coding agents.

\noindent\textbf{Benchmarks for Coding and Game Agents.}
Coding benchmarks differ in the scope of the artifact and the environment in which it is evaluated: programs satisfying input--output requirements \citep{chen2021humaneval,austin2021mbpp,hendrycks2021apps,liu2023evalplus,jain2025livecodebench}, repository-level issue resolution and the interfaces through which agents edit and execute code \citep{jimenez2024swebench,yang2024sweagent,merrill2026terminalbench}, and interaction in heterogeneous environments, websites, desktops, and embodied control \citep{liu2024agentbench,zhou2024webarena,xie2024osworld,fu2026capx}.
For games, From Code to Play, ProxyWar, and CodeClash are the closest precedents, evaluating program search, testing and repair, and repeated codebase editing under competition \citep{eberhardinger2025codeplay,peng2026proxywar,yang2025codeclash}.
Coding agents also develop the games themselves, with methods for generation from specifications~\citep{jiang2026opengame,zhou2026gamecreationagents} and refinement through playtesting~\citep{huang2026play2code,chen2026recursivegamecreator,wu2026rsigame}, benchmarks for development~\citep{chi2026gamedevbench,luo2026gamecraftbench,zhang2026gameasgbench,lee2026a2zgamespec,dong2026cegamebench} and replication~\citep{qiao2026gamereplica}, and generated environments for evaluating and training game agents~\citep{aggarwal2026gymanythingrt,liu2026script2play}.
Game agent evaluation instead builds on established platforms \citep{bellemare2013ale,kempka2016vizdoom,guss2019minerl,cote2019textworld,hausknecht2020jericho} and recent language-model suites \citep{paglieri2025balrog,hu2025lmgamebench,park2026orak}, whose conditions are essential to interpreting reported performance: real-time versus paused execution \citep{zhang2026videogamebench}, state-based verification \citep{ouyang2026gameworld,zhang2026pokegym} versus model-judged milestones \citep{zhang2026gameverse}, and improvement across attempts versus transfer to held-out variants \citep{lin2026omnigamearena}.
Coding4Games combines autonomous program development with independent testing of frozen programs, measuring performance, reliability, and cost under matched action and timing conditions.

\noindent\textbf{Self-Improving Agents.}
Self-improvement methods differ in what changes and what persists across tasks: program libraries learned together with a search-guiding model \citep{ellis2020dreamcoder}, parameters updated from accepted self-generated rationales \citep{zelikman2022star}, outputs or verbal feedback refined in context \citep{madaan2023selfrefine,shinn2023reflexion}, and reusable tools and skills accumulated with fixed language-model weights \citep{schick2023toolformer,cai2024toolmakers,zhou2026mementoskills}.
Closer to our setting, one line searches over executable policies and agent implementations \citep{kumar2026digitalredqueen,hu2025adas}, and recursive variants additionally modify the improvement machinery itself \citep{zelikman2024stop,yin2025godelagent,zhang2026darwin}---program-level modification that should be distinguished from changing the underlying model.
Coding4Games evaluates the reliability and cost of the resulting game-playing programs without prescribing a self-improvement algorithm.
\begin{table*}[t]
\centering
\newcommand{\overviewgame}[2]{\raisebox{-0.05em}{#1}\nobreak\hspace{0.3em}#2}
\footnotesize
\setlength{\tabcolsep}{5pt}
\renewcommand{\arraystretch}{1.12}
\caption{\textbf{Game categories.} The working suite groups 15 games by their principal interaction demands. Task objectives and scoring criteria appear in the Game Tasks and Scoring section of the supplementary material.}
\vspace{-1em}
\label{tab:game-tasks}
\begin{tabularx}{\linewidth}{@{}>{\raggedright\arraybackslash}p{0.22\linewidth}>{\raggedright\arraybackslash}X@{}}
\toprule
\rowcolor{tableheaderblue}
\textbf{Category} & \textbf{Games} \\
\midrule
Combat \& Action Timing & \overviewgame{\includegraphics[height=0.9em]{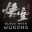}}{Black Myth: Wukong (BMW)}; \overviewgame{\includegraphics[height=0.9em]{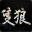}}{Sekiro: Shadows Die Twice (SEK)}; \overviewgame{\includegraphics[height=0.9em]{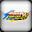}}{The King of Fighters '98 (KOF98)} \\
\addlinespace[4pt]
Strategy \& Resource Planning & \overviewgame{\includegraphics[height=0.9em]{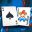}}{Balatro (BAL)}; \overviewgame{\includegraphics[height=0.9em]{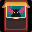}}{Slay the Spire (StS)}; \overviewgame{\includegraphics[height=0.9em]{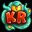}}{Kingdom Rush 5: Alliance (KR5)} \\
\addlinespace[4pt]
3D Sports \& Precision Control & \overviewgame{\includegraphics[height=0.9em]{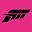}}{Forza Horizon 5 (FH5)}; \overviewgame{\includegraphics[height=0.9em]{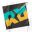}}{Riders Republic (RR)}; \overviewgame{\includegraphics[height=0.9em]{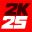}}{NBA 2K25 (NBA25)} \\
\addlinespace[4pt]
Movement \& Timing & \overviewgame{\includegraphics[height=0.9em]{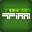}}{TrackMania Nations Forever (TM)}; \overviewgame{\includegraphics[height=0.9em]{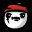}}{Brotato (BRO)}; \overviewgame{\includegraphics[height=0.9em]{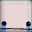}}{BattleBlock Theater (BBT)} \\
\addlinespace[4pt]
Multi-step Task Execution & \overviewgame{\includegraphics[height=0.9em]{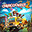}}{Overcooked! 2 (OC2)}; \overviewgame{\includegraphics[height=0.9em]{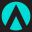}}{Aimlab (AIM)}; \overviewgame{\includegraphics[height=0.9em]{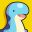}}{Palworld (PAL)} \\
\bottomrule
\end{tabularx}
\vspace{-1em}
\end{table*}

\section{Coding4Games Benchmark}
\label{sec:benchmark}

Coding4Games investigates whether modern coding agents can play video games by developing standalone executable programs. For each evaluation task, we specify a game, an objective, a starting scenario, and an interaction budget. 
Evaluating this capability requires both an environment for developing programs through game interaction and a protocol for testing the resulting programs independently. We therefore introduce a framework that connects a sandboxed development environment to a gym-like game server across a diverse set of games while keeping program development separate from program evaluation. Below, we present the evaluation framework (Sec.~\ref{sec:evaluation-framework}), the games and their tasks (Sec.~\ref{sec:games-tasks}), and the measures used to score them (Sec.~\ref{sec:evaluation-measures}).

\subsection{Evaluation framework}
\label{sec:evaluation-framework}

\begin{figure*}
    \centering
    \includegraphics[width=\linewidth]{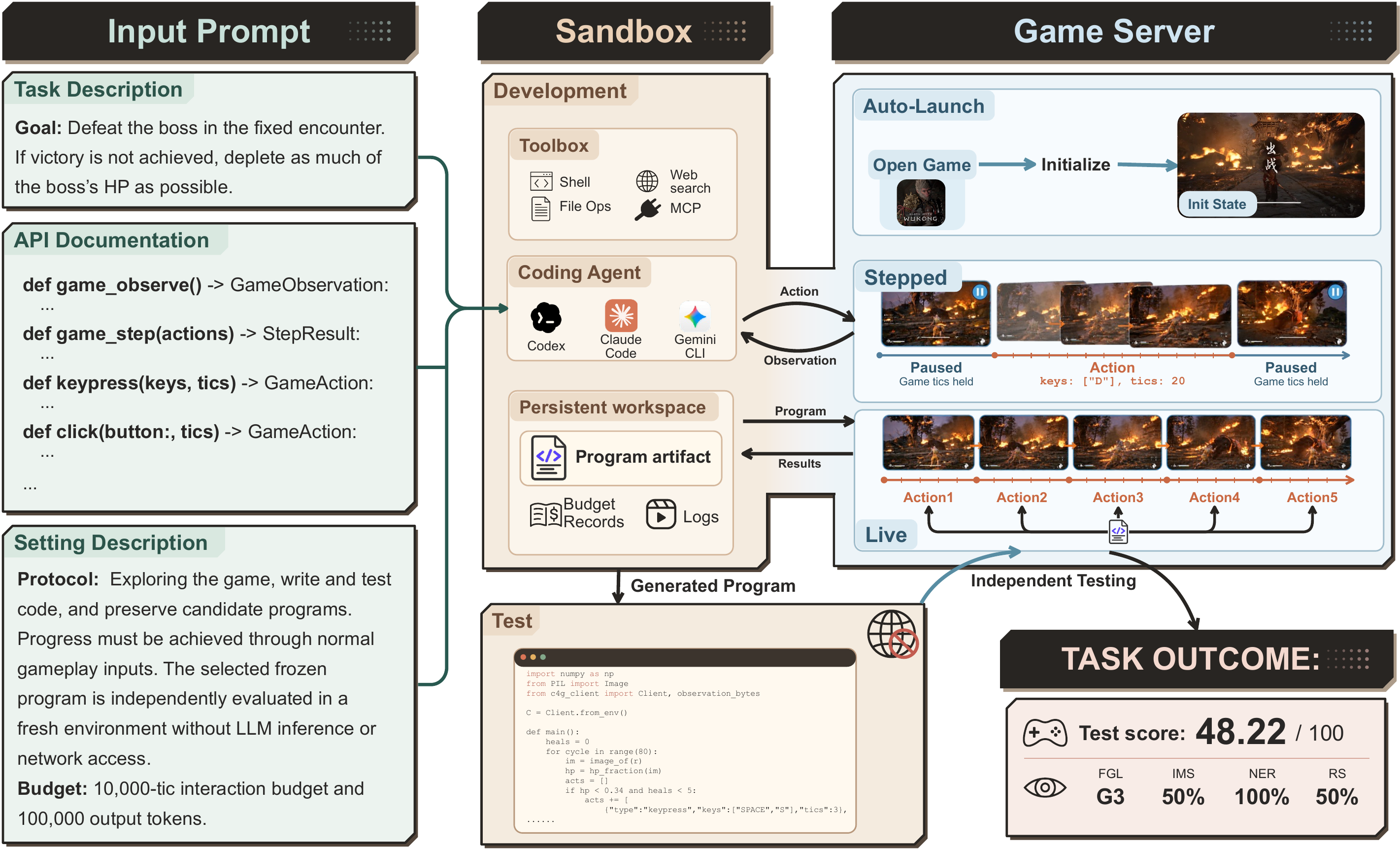}
    \vspace{-1em}
    \caption{\textbf{Evaluation framework.} The game server provides a common interface to heterogeneous games. Coding agents develop and select programs in a sandbox; the frozen programs are then evaluated through independent rollouts with host-controlled initialization, budgets, and scoring.}
    \label{fig:benchmark-framework}
    \vspace{-1em}

\end{figure*}

As shown in Fig.~\ref{fig:benchmark-framework}, the framework makes heterogeneous games accessible to coding agents and evaluates the programs they produce under independent, consistent conditions. Given the input prompt of task description, API documentation, and setting description, the coding agent combines a game server for standardized interaction, a Development stage that supports autonomous software development, and an isolated Test stage for evaluating frozen programs. Below, we introduce the main components of the evaluation framework.

\paragraph{Game Server.}
Games differ in how they start, expose observations, and accept inputs, making them difficult to access consistently from a coding environment. The game server wraps each task with a shared, gym-like interface for starting episodes, returning game screenshots, and applying gameplay inputs. It also tracks elapsed game tics and computes task-specific scores from gameplay evidence independently of the coding agent. Keeping initialization, game-tic accounting, and scoring under host control separates gameplay from measurement and supports fair, consistent evaluation across submitted programs.

\paragraph{Development Stage.}
The Development stage gives a coding agent one persistent sandbox for each development run, in which it can use a real coding CLI to write, execute, and debug programs~\citep{openai2025codexcli,anthropic2025claudecode}. Keeping this sandbox across iterations preserves the agent's tools and intermediate work and reduces variation from changes in the development environment. The sandbox is configured to block direct access to game implementation files and the host-side scorer; game observation and control instead pass through the game server's MCP interface~\citep{anthropic2024mcp}. The agent can repeatedly interact with the game, test and revise candidates, and select a program artifact, $P_{\mathrm{sel}}$; it may influence game outcomes through normal gameplay inputs, not by editing game files or scores. The host separately records model-token use and game tics, enforcing a token budget, a cumulative game-tic budget, and a per-rollout game-tic limit that bound further interaction and determine when development must stop. Agent and game interactions, candidate tests, and resource use are logged as an auditable development trajectory. Together, the stable sandbox and recorded trajectory support a reproducible experimental procedure without requiring identical gameplay outcomes.

\paragraph{Test Stage.}
The Test stage evaluates the selected frozen program in a fresh environment, separate from development~\citep{jimenez2024swebench,peng2026proxywar}. The program runs without model inference or public network access; its code stays fixed, although its internal state can change during play. Each test rollout follows the same per-rollout game-tic limit as development. Before each rollout, the host prepares the designated scenario using recorded initialization actions. After the rollout, the host-side scorer computes a task-specific score from gameplay evidence. Repeated rollouts estimate the program's performance and variability under the same task conditions.

\subsection{Games and Tasks}
\label{sec:games-tasks}

Table~\ref{tab:game-tasks} lists the games and their associated capability requirements for game-play program.
We select games to examine whether coding agents can use interaction to develop standalone game-playing programs under diverse control and planning demands. Our proposed suite comprises 15 tasks, one per game, across five categories: \textit{Combat \& Action Timing}, \textit{Strategy \& Resource Planning}, \textit{3D Sports \& Precision Control}, \textit{Movement \& Timing}, and \textit{Multi-step Task Execution}.

These tasks require different combinations of responding to game feedback, timing precise inputs, tracking resources and persistent state, and coordinating actions across stages. Each task defines a starting scenario, an objective, and an interaction budget to support repeated evaluation under consistent conditions. Each task also defines a quantitative episode score that the host computes from recorded gameplay evidence, independently of the agent's self-assessment. Following prior work~\citep{ouyang2026gameworld,zhang2026gameverse}, these task-specific scores capture partial progress beyond binary success or failure. Depending on the objective, they measure boss-health depletion, waves or floors reached, distance traveled, or orders completed, with bonuses and penalties for outcomes such as victory or wasted resources. During development, the host returns the score after each episode to guide program comparison and revision. During Test, the same scoring rule measures the performance of the selected frozen program across repeated episodes.

\begin{table*}
\providecolor{missing}{HTML}{A5ADB8}
\providecolor{tablerankwarm}{HTML}{DC9559}
\providecommand{\modelicon}[1]{\raisebox{-0.12em}{\ifnum\pdfstrcmp{#1}{kimi-color}=0\relax{\setlength{\fboxsep}{0.08em}\colorbox{black}{\includegraphics[width=0.84em,height=0.84em,keepaspectratio]{Tab/main_results/icons/#1.png}}}\else\includegraphics[width=1em,height=1em,keepaspectratio]{Tab/main_results/icons/#1.png}\fi}\hspace{0.3em}}
\centering
\footnotesize
\setlength{\tabcolsep}{1pt}
\renewcommand{\arraystretch}{1.22}
\caption{\textbf{Main results on Coding4Games.} We compare coding agents, GUI agents, and a human player using mean game scores, normalized to 0--100 within each game across all available entries. Avg. is the arithmetic mean of the 15 normalized game scores. The top 3 rows report each game's cumulative game-tic budget, per-rollout game-tic limit, and token budget. Game names follow Tab.~\ref{tab:game-tasks}; CC and GC denote Claude Code and Gemini CLI.}
\vspace{-1em}
\label{tab:main-results}
\setlength{\aboverulesep}{0pt}
\setlength{\belowrulesep}{0pt}
\resizebox{\textwidth}{!}{%
\begin{tabular}{@{}l@{\hspace{5pt}}>{\centering\arraybackslash}m{29pt}*{15}{c}|cc@{}}
\toprule
\rowcolor{tableheaderblue}
\textbf{Model} & \textbf{Harness} & \textbf{BAL} & \textbf{FH5} & \textbf{StS} & \textbf{BMW} & \textbf{BRO} & \textbf{NBA25} & \textbf{RR} & \textbf{KOF98} & \textbf{TM} & \textbf{KR5} & \textbf{OC2} & \textbf{SEK} & \textbf{AIM} & \textbf{PAL} & \textbf{BBT} & \textbf{Avg.} $\uparrow$ & \textbf{Rank} \\
\midrule
\textit{Budget (all)} &  & 10k & 10k & 10k & 10k & 10k & 10k & 10k & 10k & 10k & 20k & 10k & 10k & 1k & 10k & 10k & &  \\
\textit{Budget (rollout)} &  & 1k & 1k & 1k & 1k & 1k & 1k & 1k & 1k & 1k & 10k & 1k & 5k & 320 & 1.2k & 3k & &  \\
\textit{Budget (token)} &  & 20M & 20M & 20M & 20M & 20M & 20M & 20M & 10M & 1M & 10M & 10M & 50M & 5M & 10M & 10M & &  \\
\midrule
Human Player & \textcolor{missing}{\textemdash} & 100.0 & 99.5 & 100.0 & 100.0 & 100.0 & 100.0 & 77.7 & 100.0 & 100.0 & 100.0 & 100.0 & 100.0 & 80.2 & 100.0 & 100.0 & 97.2 & \textcolor{missing}{\textemdash} \\
\rowcolor{tablegroupgray}\multicolumn{17}{@{}l|}{\textcolor{tabletextgray}{\bfseries\itshape GUI Agent}} & \multicolumn{2}{c@{}}{} \\
\modelicon{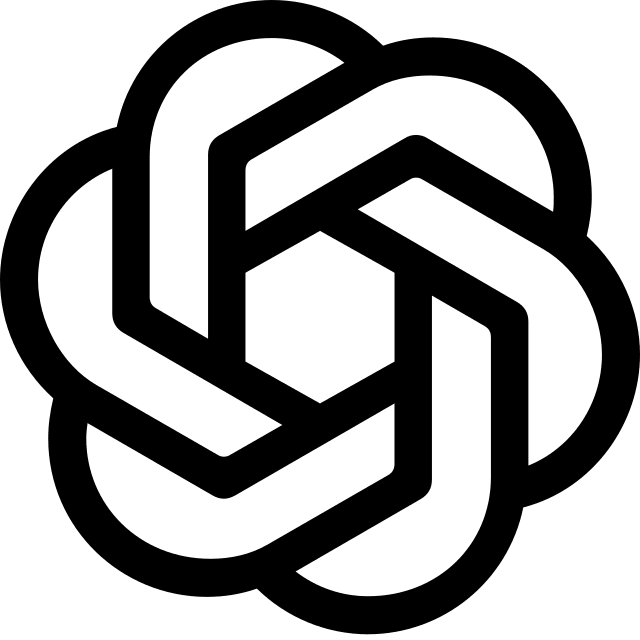}GPT-5.6-Sol & Codex & 4.8 & 70.7 & 8.3 & 29.5 & 13.7 & 80.7 & 5.4 & 32.8 & 16.3 & 1.6 & 0.0 & 11.2 & 22.0 & 85.1 & 14.0 & 26.4 & \textcolor{missing}{\textemdash} \\
\modelicon{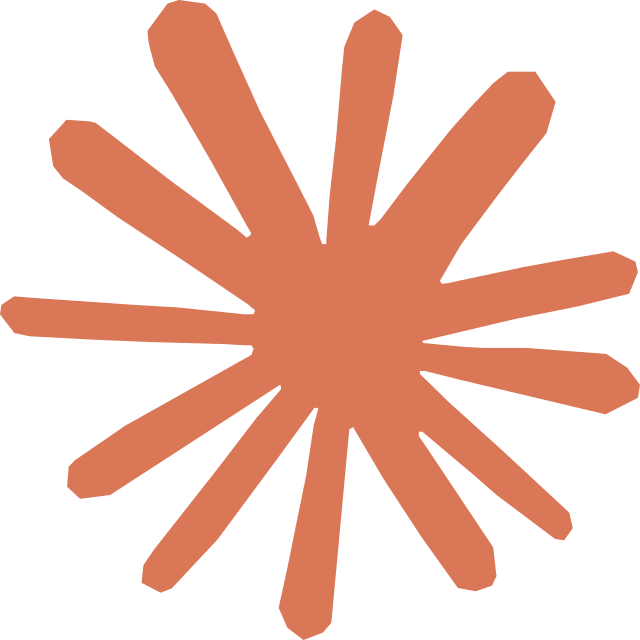}Claude-Sonnet-5 & CC & 4.0 & 100.0 & 66.8 & 12.3 & 93.3 & 38.7 & 18.6 & 8.2 & 0.0 & 1.8 & 0.0 & 0.0 & 24.9 & 30.0 & 3.7 & 26.8 & \textcolor{missing}{\textemdash} \\
\rowcolor{tablegroupgray}\multicolumn{17}{@{}l|}{\textcolor{tabletextgray}{\bfseries\itshape Coding Agent}} & \multicolumn{2}{c@{}}{} \\
\multicolumn{17}{@{}l|}{\textcolor{tabletextgray}{\bfseries Closed}} & \multicolumn{2}{c@{}}{} \\

\modelicon{openai}GPT-6-Astra & Codex & \cellcolor{tablerankwarm}\textcolor{black}{100.0} & \cellcolor{tablerankwarm}\textcolor{black}{100.0} & \cellcolor{tablerankwarm}\textcolor{black}{100.0} & 37.8 & \cellcolor{tablerankwarm}\textcolor{black}{96.0} & 45.2 & 26.2 & \cellcolor{tablerankwarm!55!white}\textcolor{black}{80.9} & \cellcolor{tablerankwarm}\textcolor{black}{51.6} & 27.3 & 0.0 & 1.9 & \cellcolor{tablerankwarm!25!white}\textcolor{black}{64.4} & 50.4 & \cellcolor{tablerankwarm!25!white}\textcolor{black}{7.0} & 52.6 & \raisebox{-0.3em}{\includegraphics[height=1.6em]{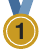}} \\
\modelicon{openai}GPT-5.6-Sol & Codex & 14.8 & 10.0 & 0.0 & 39.6 & 5.3 & \cellcolor{tablerankwarm!25!white}\textcolor{black}{58.1} & \cellcolor{tablerankwarm!25!white}\textcolor{black}{47.3} & 60.1 & 0.9 & \cellcolor{tablerankwarm}\textcolor{black}{74.2} & \cellcolor{tablerankwarm!55!white}\textcolor{black}{3.0} & \cellcolor{tablerankwarm}\textcolor{black}{13.7} & 41.1 & 15.0 & 7.0 & 26.0 & 6 \\
\modelicon{openai}GPT-5.6-Terra & Codex & 15.4 & \cellcolor{tablerankwarm!25!white}\textcolor{black}{49.1} & 0.0 & \cellcolor{tablerankwarm}\textcolor{black}{73.2} & 37.1 & 0.0 & \cellcolor{tablerankwarm!55!white}\textcolor{black}{53.1} & 54.6 & 1.9 & 28.9 & 0.0 & 7.2 & 52.4 & 6.5 & 7.0 & 25.8 & 7 \\
\modelicon{claude-color}Claude-Opus-5.5 & CC & \cellcolor{tablerankwarm!55!white}\textcolor{black}{50.8} & \cellcolor{tablerankwarm!55!white}\textcolor{black}{84.0} & 42.1 & 16.6 & 84.3 & \cellcolor{tablerankwarm}\textcolor{black}{77.4} & 22.5 & 37.0 & \cellcolor{tablerankwarm!55!white}\textcolor{black}{39.8} & 25.8 & \cellcolor{tablerankwarm!25!white}\textcolor{black}{1.5} & \cellcolor{tablerankwarm!55!white}\textcolor{black}{10.3} & \cellcolor{tablerankwarm}\textcolor{black}{100.0} & \cellcolor{tablerankwarm!25!white}\textcolor{black}{53.9} & \cellcolor{tablerankwarm!55!white}\textcolor{black}{14.1} & 44.0 & \raisebox{-0.3em}{\includegraphics[height=1.6em]{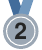}} \\
\modelicon{claude-color}Claude-Opus-5 & CC & \cellcolor{tablerankwarm!25!white}\textcolor{black}{29.4} & 32.4 & \cellcolor{tablerankwarm!55!white}\textcolor{black}{57.9} & 38.1 & \cellcolor{tablerankwarm!25!white}\textcolor{black}{89.1} & \cellcolor{tablerankwarm!25!white}\textcolor{black}{58.1} & 11.1 & 16.4 & 0.8 & 28.3 & 0.0 & 2.9 & \cellcolor{tablerankwarm!55!white}\textcolor{black}{94.7} & \cellcolor{tablerankwarm!55!white}\textcolor{black}{68.0} & \cellcolor{tablerankwarm}\textcolor{black}{37.7} & 37.7 & \raisebox{-0.3em}{\includegraphics[height=1.6em]{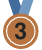}} \\
\modelicon{claude-color}Claude-Sonnet-5 & CC & 9.0 & 35.9 & 0.0 & 12.1 & 0.7 & 38.7 & 0.0 & \cellcolor{tablerankwarm!25!white}\textcolor{black}{62.8} & 13.5 & 15.4 & 0.0 & 2.7 & 45.3 & 24.3 & 7.0 & 17.8 & 8 \\
\modelicon{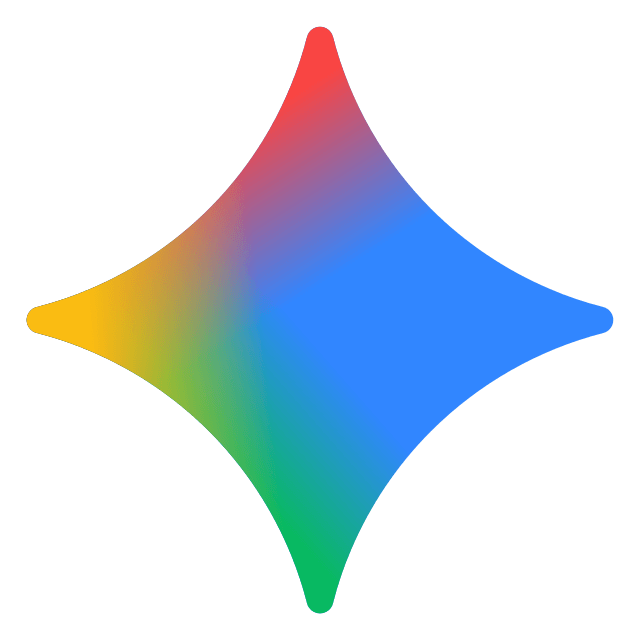}Gemini-3.8-Flash & GC & 18.8 & 34.2 & \cellcolor{tablerankwarm!25!white}\textcolor{black}{42.2} & 3.7 & 30.3 & 22.6 & 17.9 & \cellcolor{tablerankwarm}\textcolor{black}{98.4} & 18.1 & 27.3 & \cellcolor{tablerankwarm!25!white}\textcolor{black}{1.5} & \cellcolor{tablerankwarm!25!white}\textcolor{black}{10.0} & 41.6 & \cellcolor{tablerankwarm}\textcolor{black}{89.0} & 7.0 & 30.8 & 5 \\
\modelicon{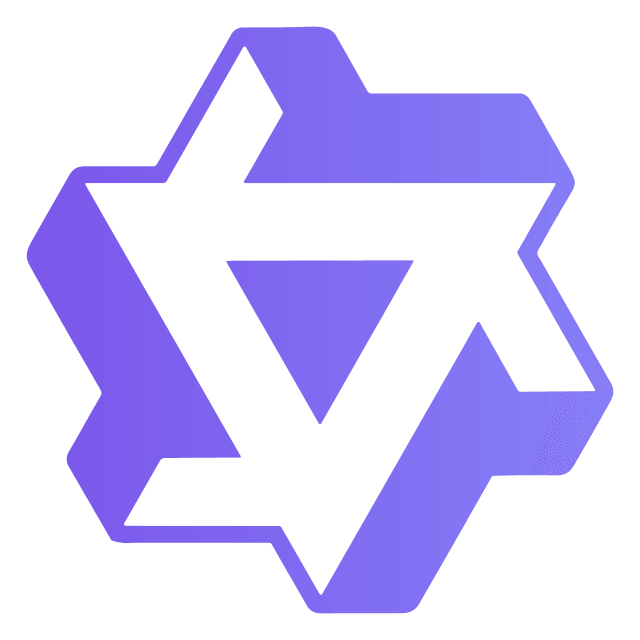}Qwen3.8-Flash & Codex & 0.0 & 20.8 & 0.0 & \cellcolor{tablerankwarm!55!white}\textcolor{black}{65.4} & \cellcolor{tablerankwarm!55!white}\textcolor{black}{95.1} & 32.3 & \cellcolor{tablerankwarm}\textcolor{black}{100.0} & 19.1 & 20.6 & 19.6 & 0.0 & 5.5 & 51.3 & 41.1 & 3.7 & 31.6 & 4 \\
\multicolumn{17}{@{}l|}{\textcolor{tabletextgray}{\bfseries Open}} & \multicolumn{2}{c@{}}{} \\
\modelicon{qwen-color}Qwen3.8-27B & Codex & 16.7 & 10.0 & 21.0 & 0.0 & 1.2 & \cellcolor{tablerankwarm!25!white}\textcolor{black}{58.1} & 14.1 & 5.5 & 4.6 & \cellcolor{tablerankwarm!25!white}\textcolor{black}{28.9} & 0.0 & 0.0 & 22.6 & 5.1 & 0.0 & 12.5 & 11 \\
\modelicon{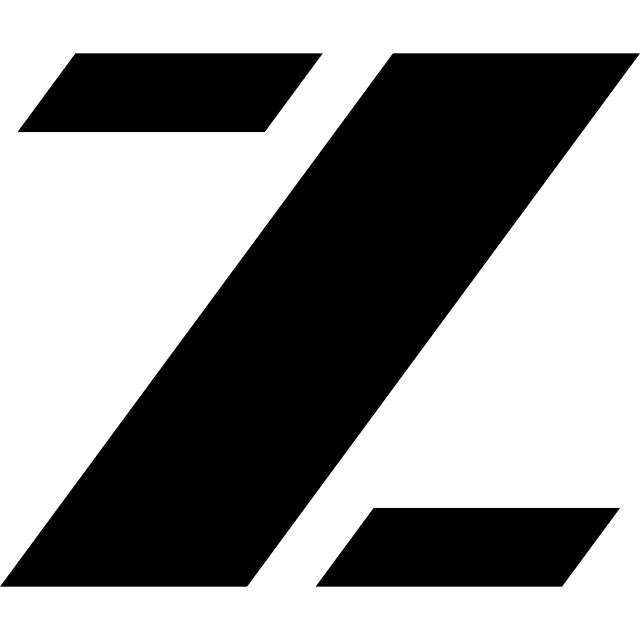}GLM-5.3-Flash & Codex & 14.5 & 19.0 & 0.0 & 5.1 & 1.3 & \cellcolor{tablerankwarm!55!white}\textcolor{black}{71.0} & 1.3 & 0.0 & \cellcolor{tablerankwarm!25!white}\textcolor{black}{34.0} & 0.0 & \cellcolor{tablerankwarm}\textcolor{black}{7.4} & 1.6 & 40.6 & 0.0 & 0.0 & 13.1 & 10 \\
\mbox{\modelicon{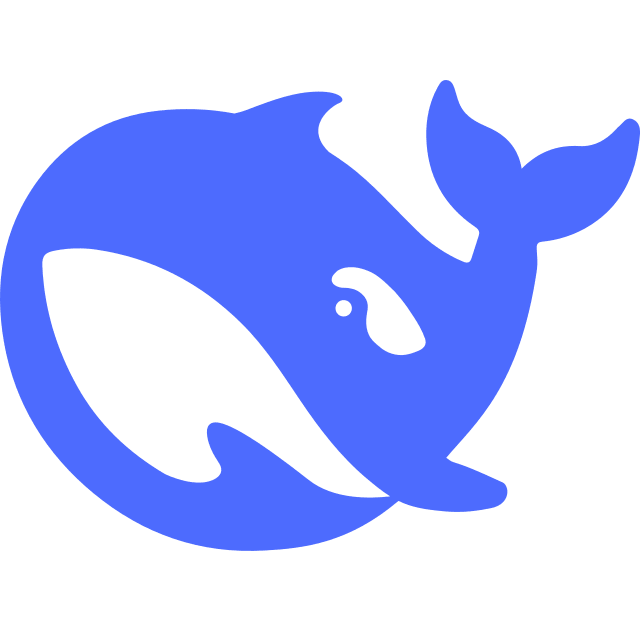}DeepSeek-V4.1-Flash} & Codex & 0.0 & 0.0 & 36.8 & 10.3 & 28.5 & 6.4 & 16.7 & 10.9 & 33.9 &\cellcolor{tablerankwarm!55!white}\textcolor{black}{30.4} & \cellcolor{tablerankwarm!25!white}\textcolor{black}{1.5} & 0.7 & 32.3 & 40.1 & 7.0 & 17.0 & 9 \\
\modelicon{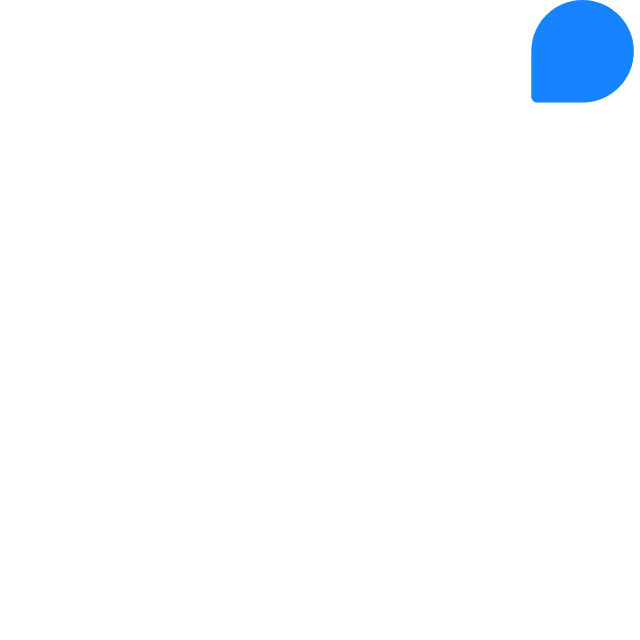}Kimi-K3 & Codex & 0.0 & 15.0 & 0.0 & \cellcolor{tablerankwarm!25!white}\textcolor{black}{47.2} & 0.0 & 54.8 & 32.4 & 16.4 & 1.2 & 1.8 & \cellcolor{tablerankwarm!25!white}\textcolor{black}{1.5} & 9.7 & 0.0 & 0.8 & 0.0 & 12.1 & 12 \\
\bottomrule
\end{tabular}%
}
\vspace{4pt}
\end{table*}

\subsection{Evaluation measures}
\label{sec:evaluation-measures}
We design these measures to understand what coding agents produce from game interaction, how they develop it, and at what cost. Independent game scores show how well the selected program plays; the other measures examine whether it uses game state, how deeply the agent revises it, whether revisions address observed problems, and the resources required for development and deployment. Together, they distinguish programs with similar game scores but different decision mechanisms, development trajectories, and costs.

\paragraph{Game performance.}
For each model--game pair, we evaluate the program selected during development in $N$ independent test episodes and use the mean episode score as its game performance:
\begin{equation}
    \bar{s}=\frac{1}{N}\sum_{k=1}^{N}s_k,
    \label{eq:program-performance}
\end{equation}
where $s_k$ is the score in test episode $k$. Repeated testing reduces sensitivity to randomness in the game and variation in the program's behavior, either of which can change the outcome of a single episode. We also report variation across episodes.

\paragraph{Dependence on game state.}
We propose \emph{Final Grounding Level} (FGL) to describe how the selected program uses game information to choose actions: \textbf{G0}, fixed actions; \textbf{G1}, actions based on time or counters; \textbf{G2}, reactions to simple observed signals; \textbf{G3}, decisions based on task-relevant game state; and \textbf{G4}, state-based decisions with memory, planning, and failure recovery. We use these levels to study how the agent encodes its understanding of the game in the program and turns that understanding into gameplay control.

\paragraph{Modification depth.}
We propose \emph{Iterative Modification Score} (IMS) to measure the depth of changes between successive self-tested program versions. Each change receives a rating $d_i$ from 0 to 4: no change, parameter adjustment, local patch, component change, or strategy shift. For $N$ changes, IMS is the average rating scaled to 0--100:
\begin{equation}
    \mathrm{IMS}=\frac{100}{4N}\sum_{i=1}^{N}d_i.
    \label{eq:ims}
\end{equation}
 A higher value means deeper changes, not necessarily better results. We compare each tested score and resource use to see how performance evolves during development~\citep{lin2026omnigamearena}.

\paragraph{Problem resolution.}
For each problem identified before a modification, we check whether the next self-test reaches a situation where that problem can be assessed. Across $M$ problem--modification pairs, let $e_j=1$ if pair $j$ receives such a test and 0 otherwise. Let $r_j$ be 0 if the problem remains, 0.5 if partly resolved, and 1 if resolved. We measure how often modifications can be assessed (NER) and how fully the assessed problems are resolved (RS):
\begin{equation}
    \mathrm{NER}=\frac{\sum_{j=1}^{M}e_j}{M},
    \qquad
    \mathrm{RS}=\frac{\sum_{j=1}^{M}e_jr_j}{\sum_{j=1}^{M}e_j}.
    \label{eq:resolution}
\end{equation}
NER is undefined when $M=0$; RS is undefined if no pair receives a relevant test. Together they distinguish untested changes from observed problem resolution.

\section{Experiments}
\label{sec:experiments}

\newsavebox{\processtablebox}
\newsavebox{\pairedresultstablebox}
\newlength{\processbodyheight}
\newlength{\pairedbodyheight}
\newcommand{\alignedprocessbody}[1]{%
  \vbox to \processbodyheight{%
    \hbox{\usebox{#1}}%
    \vfil
    \vskip\aboverulesep
    \hrule height \heavyrulewidth
  }%
}
\begin{table*}
\providecommand{\modelicon}[1]{\raisebox{-0.12em}{\ifnum\pdfstrcmp{#1}{kimi-color}=0\relax{\setlength{\fboxsep}{0.08em}\colorbox{black}{\includegraphics[width=0.84em,height=0.84em,keepaspectratio]{Tab/main_results/icons/#1.png}}}\else\includegraphics[width=1em,height=1em,keepaspectratio]{Tab/main_results/icons/#1.png}\fi}\hspace{0.3em}}
\newcommand{\pairedresultscaption}{%
\caption{\textbf{GUI agents vs. programs.}
Per-model score gaps (programs minus GUI).
Blue favors GUI agents, and orange favors programs. Categories follow Tab.~\ref{tab:game-tasks}.}
\label{tab:gui-code-performance}}
\newcommand{\pairedgameicon}[1]{\raisebox{-0.05em}{\includegraphics[height=0.9em]{Tab/game_tasks/icons/#1}}\nobreak\hspace{0.3em}}
\newcommand{\pairedresultsheading}[1]{%
  \raisebox{\dimexpr(\ht\strutbox-\dp\strutbox-\height+\depth)/2\relax}%
    [\dimexpr\ht\strutbox+\baselineskip/2\relax]%
    [\dimexpr\dp\strutbox+\baselineskip/2\relax]{#1}%
}
\newcommand{\pairedresultstabular}[1][0pt]{%
\setlength{\tabcolsep}{2pt}%
\renewcommand{\arraystretch}{1.0}%
\definecolor{pairwarm}{HTML}{E6A68A}%
\definecolor{paircool}{HTML}{A6BCD4}%
\begin{tabularx}{\linewidth}{@{}>{\raggedright\arraybackslash}Xl*{2}{>{\centering\arraybackslash}p{30pt}}@{}}
\toprule
\rowcolor{tableheaderblue}
\pairedresultsheading{\textbf{Game}} & \pairedresultsheading{\textbf{Category}} & \pairedresultsheading{\shortstack{\textbf{GPT-5.6-}\\\textbf{Sol}}} & \pairedresultsheading{\shortstack{\textbf{Claude-}\\\textbf{Sonnet-5}}} \\
\midrule
\pairedgameicon{steam_2379780.jpg}Balatro & Strategy & \cellcolor{pairwarm!10}$+10.0$ & \cellcolor{pairwarm!5}$+5.0$ \\[#1]
\pairedgameicon{steam_1551360.jpg}Forza Horizon 5 & Sports & \cellcolor{paircool!61}$-60.7$ & \cellcolor{paircool!64}$-64.1$ \\[#1]
\pairedgameicon{steam_646570.jpg}Slay the Spire & Strategy & \cellcolor{paircool!8}$-8.3$ & \cellcolor{paircool!67}$-66.8$ \\[#1]
\pairedgameicon{steam_2358720.jpg}Black Myth: Wukong & Combat & \cellcolor{pairwarm!10}$+10.1$ & \cellcolor{paircool!0}$-0.2$ \\[#1]
\pairedgameicon{steam_1942280.jpg}Brotato & Movement & \cellcolor{paircool!8}$-8.4$ & \cellcolor{paircool!93}$-92.6$ \\[#1]
\pairedgameicon{steam_2878980.jpg}NBA 2K25 & Sports & \cellcolor{paircool!23}$-22.6$ & $0.0$ \\[#1]
\pairedgameicon{steam_2290180.jpg}Riders Republic & Sports & \cellcolor{pairwarm!42}$+41.9$ & \cellcolor{paircool!19}$-18.6$ \\[#1]
\pairedgameicon{steam_222420.jpg}The King of Fighters '98 & Combat & \cellcolor{pairwarm!27}$+27.3$ & \cellcolor{pairwarm!55}$+54.6$ \\[#1]
\pairedgameicon{steam_11020.jpg}TrackMania Nations Forever & Movement & \cellcolor{paircool!15}$-15.4$ & \cellcolor{pairwarm!14}$+13.5$ \\[#1]
\pairedgameicon{steam_2849080.jpg}Kingdom Rush 5: Alliance & Strategy & \cellcolor{pairwarm!73}$+72.6$ & \cellcolor{pairwarm!14}$+13.6$ \\[#1]
\pairedgameicon{steam_728880.jpg}Overcooked! 2 & Multi-step & \cellcolor{pairwarm!3}$+3.0$ & $0.0$ \\[#1]
\pairedgameicon{steam_814380.jpg}Sekiro: Shadows Die Twice & Combat & \cellcolor{pairwarm!3}$+2.5$ & \cellcolor{pairwarm!3}$+2.7$ \\[#1]
\pairedgameicon{steam_714010.jpg}Aimlab & Multi-step & \cellcolor{pairwarm!19}$+19.1$ & \cellcolor{pairwarm!20}$+20.4$ \\[#1]
\pairedgameicon{steam_1623730.jpg}Palworld & Multi-step & \cellcolor{paircool!70}$-70.1$ & \cellcolor{paircool!6}$-5.7$ \\[#1]
\pairedgameicon{steam_238460.jpg}BattleBlock Theater & Movement & \cellcolor{paircool!7}$-7.0$ & \cellcolor{pairwarm!3}$+3.3$ \\
\end{tabularx}}

\centering
\captionsetup{hypcap=false}
\begin{minipage}[t]{0.56\textwidth}
\vspace{0pt}
\caption{\textbf{Process evaluation of coding agents.} G0--G4 show the percentage of final programs at each
grounding level. IMS, NER, and RS summarize the development process
on a 0--100 scale. See Sec.~\ref{sec:evaluation-measures} for metric definitions.}
\label{tab:process-evaluation}
\end{minipage}\hfill
\begin{minipage}[t]{0.41\textwidth}
\vspace{0pt}
\pairedresultscaption
\end{minipage}\par
\vspace{-1em}
\newcommand{\saveprocessresults}[1]{%
\begin{lrbox}{\processtablebox}
\begin{minipage}{0.56\textwidth}
\scriptsize
\setlength{\tabcolsep}{1.5pt}
\renewcommand{\arraystretch}{1.0}
\begin{tabularx}{\linewidth}{@{}l*{8}{>{\raggedleft\arraybackslash}X}@{}}
\toprule
\rowcolor{tableheaderblue}
\pairedresultsheading{\textbf{Model}} & \pairedresultsheading{\textbf{G0}} & \pairedresultsheading{\textbf{G1}} & \pairedresultsheading{\textbf{G2}} & \pairedresultsheading{\textbf{G3}} & \pairedresultsheading{\textbf{G4}} & \pairedresultsheading{\textbf{IMS}} & \pairedresultsheading{\textbf{NER}} & \pairedresultsheading{\textbf{RS}} \\
\midrule
\rowcolor{tablegroupgray}\multicolumn{9}{l}{\textcolor{tabletextgray}{\bfseries\itshape Closed}} \\
\modelicon{openai}GPT-6-Astra & 13.3 & 0.0 & 26.7 & 60.0 & 0.0 & 44.5 & 87.0 & 62.5 \\[#1]
\modelicon{openai}GPT-5.6-Sol & 9.1 & 63.6 & 27.3 & 0.0 & 0.0 & 64.3 & 100.0 & 41.7 \\[#1]
\modelicon{openai}GPT-5.6-Terra & 18.2 & 63.6 & 9.1 & 9.1 & 0.0 & 62.5 & 100.0 & 62.5 \\[#1]
\modelicon{claude-color}Claude-Opus-5.5 & 20.0 & 0.0 & 13.3 & 46.7 & 20.0 & 54.3 & 93.3 & 71.4 \\[#1]
\modelicon{claude-color}Claude-Opus-5 & 0.0 & 10.0 & 20.0 & 20.0 & 50.0 & 53.6 & 93.0 & 42.5 \\[#1]
\modelicon{claude-color}Claude-Sonnet-5 & 0.0 & 18.2 & 36.4 & 27.3 & 18.2 & 50.0 & 100.0 & 83.3 \\[#1]
\modelicon{gemini-color}Gemini-3.8-flash & 9.1 & 18.2 & 18.2 & 36.4 & 18.2 & 67.5 & 94.4 & 58.8 \\[#1]
\modelicon{qwen-color}Qwen3.8-flash & 20.0 & 0.0 & 10.0 & 20.0 & 50.0 & 56.4 & 85.3 & 44.8 \\[#1]
\midrule
\rowcolor{tablegroupgray}\multicolumn{9}{l}{\textcolor{tabletextgray}{\bfseries\itshape Open}} \\
\modelicon{qwen-color}Qwen3.8-27B & 0.0 & 0.0 & 20.0 & 50.0 & 30.0 & 50.7 & 82.9 & 50.0 \\[#1]
\modelicon{zai}GLM-5.3-Flash & 30.0 & 30.0 & 10.0 & 30.0 & 0.0 & 56.8 & 100.0 & 35.7 \\[#1]
\modelicon{deepseek-color}deepseek-v4.1-flash & 0.0 & 0.0 & 10.0 & 50.0 & 30.0 & 67.0 & 80.0 & 56.2 \\[#1]
\modelicon{kimi-color}Kimi-K3 & 9.1 & 9.1 & 27.3 & 18.2 & 36.4 & 57.6 & 95.0 & 42.1 \\
\end{tabularx}
\end{minipage}
\end{lrbox}%
}
\newcommand{\savepairedresults}[1]{%
  \begin{lrbox}{\pairedresultstablebox}
  \begin{minipage}{0.41\textwidth}
  \scriptsize
  \pairedresultstabular[#1]
  \end{minipage}
  \end{lrbox}%
}
\saveprocessresults{0pt}
\savepairedresults{0pt}
\setlength{\processbodyheight}{\dimexpr\ht\processtablebox+\dp\processtablebox\relax}
\setlength{\pairedbodyheight}{\dimexpr\ht\pairedresultstablebox+\dp\pairedresultstablebox\relax}
\ifdim\processbodyheight>\pairedbodyheight
  \savepairedresults{\dimexpr(\processbodyheight-\pairedbodyheight)/14\relax}
\else
  \saveprocessresults{\dimexpr(\pairedbodyheight-\processbodyheight)/11\relax}
  \setlength{\processbodyheight}{\pairedbodyheight}
\fi
\addtolength{\processbodyheight}{\dimexpr\aboverulesep+\heavyrulewidth\relax}
\noindent\alignedprocessbody{\processtablebox}\hfill
\alignedprocessbody{\pairedresultstablebox}\par
\end{table*}

Our evaluation includes 12 coding-agent models. The closed-model group comprises GPT-6-Astra~\citep{openai2026gpt6}, \mbox{GPT-5.6-Sol} and GPT-5.6-Terra~\citep{openai2026gpt56}, Claude-Opus-5.5~\citep{anthropic2026claudeopus5.5} and Claude-Opus-5~\citep{anthropic2026claudeopus5}, \mbox{Claude-Sonnet-5}~\citep{anthropic2026claudesonnet5}, Gemini-3.8-flash~\citep{googledeepmind2026gemini38flash}, and Qwen3.8-flash~\citep{qwenteam2026qwen38flash}. The open-model group comprises Qwen3.8-27B~\citep{qwenteam2026qwen3827b}, GLM-5.3-Flash~\citep{zai2026glm53flash}, deepseek-v4.1-flash~\citep{deepseekai2026deepseekv41flash}, and Kimi-K3~\citep{kimiteam2026kimik3}. We use each model's associated coding harness where available and Codex as a common harness for the remaining open models.
The human reference consists of a single player who is briefed on the basic rules and controls of each game but has not previously developed proficiency in playing it. The GUI baselines use \mbox{GPT-5.6-Sol} and \mbox{Claude-Sonnet-5}, with the game paused during agent inference.
We set token and game-tic budgets tailored to each game, with the specific values reported in Table~\ref{tab:main-results}. We run three test episodes per method and game and average the valid scores.

\begin{figure}[!t]
  \centering
  \includegraphics[height=.59\linewidth]{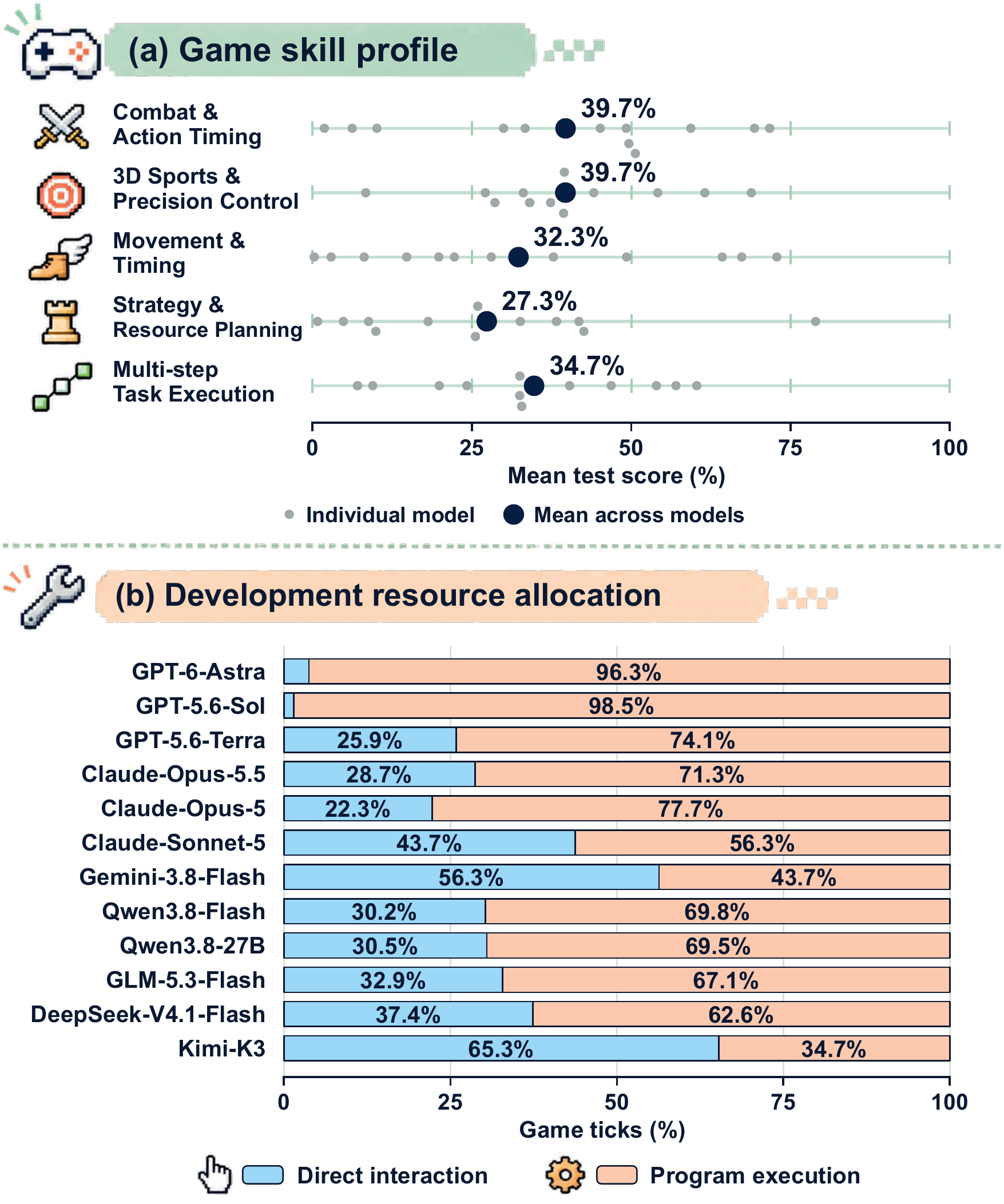}%
  \hspace{.025\linewidth}%
  \includegraphics[height=.59\linewidth]{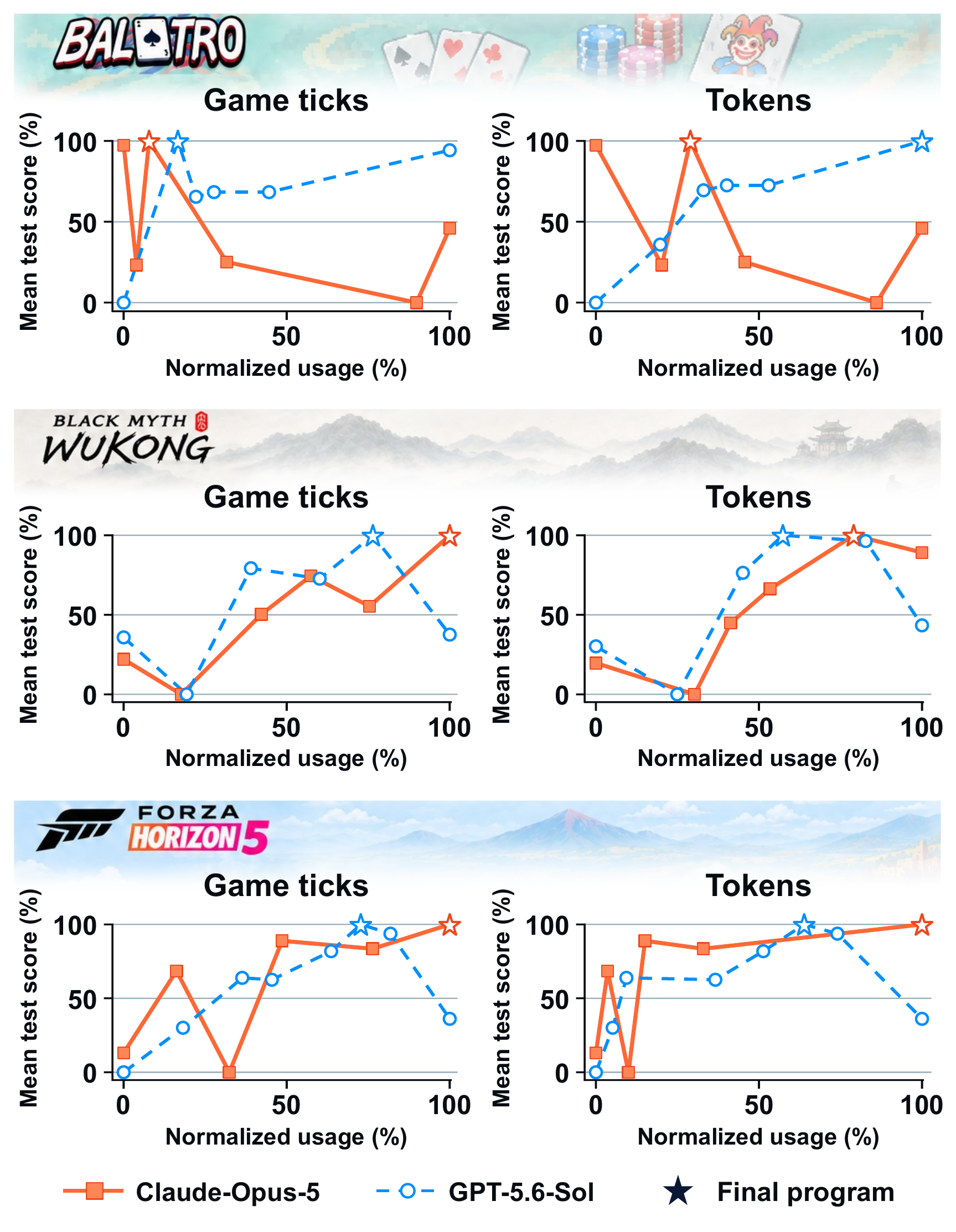}%
  \caption{\textbf{Game-category performance and development interaction modes (left).}
  (a) For each game, scores are min--max normalized to 0--1 across the 12 coding models, then averaged across games in each category.
  (b) Direct and Program game tics as fractions of their sum within one illustrative development run per model.
  \textbf{Development progress \emph{vs.}\ test performance (right).}
  With development capped at 15k game tics and 25M tokens, nonuniformly sampled programs are independently tested for each game.
  Each row shows the mean test score against game tics and tokens for one game.
  Axes are rescaled from 0 to 100\% per curve, and stars mark final selected programs.}
  \label{fig:category-tics}
  \label{fig:tics-performance}
  \label{fig:tokens-performance}
\end{figure}

\subsection{Main results}
\label{sec:main-results}

\noindent\textbf{Overall performance.}
Table~\ref{tab:main-results} shows substantial variation across models and games. Eight coding models achieve the highest coding score on at least one game: GPT-6-Astra leads on five, while Claude-Opus-5.5 leads on NBA 2K25 and Aimlab. Panel~(a) on the left of Figure~\ref{fig:category-tics} also shows different category profiles, with \mbox{GPT-5.6-Sol} strongest in Combat \& Action Timing and Claude-Opus-5.5 in 3D Sports \& Precision Control. Even within a category, no model consistently leads: GPT-6-Astra scores highest among coding agents on Balatro and Slay the Spire, whereas \mbox{GPT-5.6-Sol} leads on Kingdom Rush 5. The human player outperforms all agents on 10 of 15 games; GUI agents exceed this reference on Forza Horizon 5, and coding agents do so on Balatro, Forza Horizon 5, Slay the Spire, Riders Republic, and Aimlab. In the Computer Use Comparison, \mbox{GPT-5.6-Sol}'s frozen programs outperform its GUI baseline on eight games and underperform on seven; \mbox{Claude-Sonnet-5}'s programs win on seven, lose on six, and tie on two. Together, these results show that strong performance remains specific to particular models and games, and that frozen programs offer an alternative to online GUI control whose relative effectiveness varies across tasks.

\subsection{Understanding the performance}
\label{sec:understanding-performance}

\noindent\textbf{1. What strategies do agents implement?}
\emph{Models differ in how they use game information to control actions.} Table~\ref{tab:process-evaluation} shows that \mbox{GPT-5.6-Sol} mainly produces G1 controllers, which select actions according to timing or counters. In contrast, GPT-6-Astra more often produces G3 programs that use game state to guide decisions, while G4 is the most common level for Claude Opus 5 and Qwen3.8-flash, indicating the use of memory, planning, and recovery alongside game state. These distributions distinguish programs that largely follow predefined schedules from those that adjust decisions using feedback and accumulated information. Coding agents therefore differ in the control logic they encode, although greater reliance on game state does not by itself establish better performance.

\noindent\textbf{2. Do more complex strategies perform better?}
\emph{Greater strategy complexity does not necessarily yield higher game performance.} In Forza Horizon 5, Qwen3.8-flash's PID controller achieves a mean score of 443.7, compared with 645.7 for \mbox{Claude-Sonnet-5}'s simpler local route follower. Qwen3.8-flash's development traces record road departures and false recovery triggers, pointing to unresolved calibration issues, although their exact contribution to the score gap cannot be isolated. This gap between writing complex control code and calibrating it at runtime helps explain why stronger conventional programming performance may not transfer to standalone gameplay.

\noindent\textbf{3. Do agents resolve failures through self-testing?}
\emph{Agents can identify problems during self-testing without reliably repairing them.} Table~\ref{tab:process-evaluation} reports NER values ranging from 80.0\% to 100\%, but RS values spanning a wider range from 35.7\% to 83.3\% for identified problems targeted by modifications. \mbox{GPT-5.6-Sol} and Gemini-3.8-flash achieve NER values of 100.0\% and 94.4\%, respectively, yet their RS values are 41.7\% and 58.8\%, respectively. Obtaining relevant test feedback therefore does not ensure successful repair. Moreover, local problem resolution still requires independent testing to establish improved game performance.

\noindent\textbf{4. How do agents interact during debugging?}
\emph{Most models rely primarily on script execution to obtain feedback during development and debugging.} Panel~(b) on the left of Figure~\ref{fig:category-tics} shows that in the illustrated development runs, program execution accounts for most game tics for 10 of 12 models, reaching 98.5\% for \mbox{GPT-5.6-Sol}. Direct interaction accounts for most game tics for Gemini-3.8-Flash (56.3\%) and Kimi-K3 (65.3\%). These shares characterize development interaction, but do not distinguish exploration from targeted debugging or measure log-inspection time. Unequal run counts and budgets also limit their interpretation as stable model preferences.

\noindent\textbf{5. Where do frozen programs fail?}
\emph{Partial state representations limit how frozen programs adapt to feedback.} In Riders Republic, \mbox{GPT-5.6-Sol}'s selected program detects crashes using a red-screen threshold but executes tricks on a fixed counter-based schedule. Feedback thus supports recovery without adapting trick selection or timing to the gameplay state: detecting failures alone is insufficient to guide the actions that precede them.

\noindent\textbf{6. How do frozen programs compare with GUI agents?}
\emph{Frozen programs can match or exceed the same model's GUI agent on some games without model inference during testing.} Table~\ref{tab:gui-code-performance} compares paired results for \mbox{GPT-5.6-Sol} and \mbox{Claude-Sonnet-5}. \mbox{GPT-5.6-Sol}'s programs score higher on eight of 15 paired games and lower on seven; \mbox{Claude-Sonnet-5}'s programs score higher on seven, lower on six, and tie on two. Both models' programs outperform their GUI agents on Balatro, Kingdom Rush 5, and KOF98, spanning strategy and combat tasks. However, both GUI agents score higher on Forza Horizon 5, Slay the Spire, Brotato, and Palworld. These results show that programs produced during development can support comparable or better gameplay on a subset of tasks without continued model involvement, but they do not yet provide a general replacement for online GUI-agent decision-making.


\noindent\textbf{7. How do development budgets affect frozen-program performance?}
\emph{Both token and game-tic budgets influence performance, but their benefits vary across games.}
We examine how game-tic and token budgets affect test performance for Claude Opus 5 and \mbox{GPT-5.6-Sol} on Balatro, Black Myth: Wukong, and Forza Horizon 5 in the right panel of Fig.~\ref{fig:tics-performance}. We vary the cumulative game-tic budget from 5k to 15k in increments of 2k while fixing the token budget at 10 million, and vary the token budget from 5 to 25 million in increments of 3 million while fixing the game-tic budget at 10k. 
\section{Conclusion}
\label{sec:conclusion}

We introduced \mbox{Coding4Games}, a benchmark that asks coding agents to develop standalone programs that play 15 modern games from visual observations through keyboard-and-mouse inputs, separating autonomous development in a sandbox from independent, model-free testing of a frozen program, and evaluating the development process alongside game performance and cost. Across 12 coding agents, the best program per game outscores the strongest GUI baseline on 13 of 15 games and ties it on one. The human player outscores all agents on 10 games. Our process measures reveal a gap between self-testing and problem resolution: agents expose modifications to self-testing but often fail to resolve the failures they target, leaving programs that cover part of the gameplay conditions or rest on visual assumptions that do not hold outside the observed scenario. Turning execution feedback into revisions that generalize is, on this evidence, the central open problem for coding agents in dynamic environments.

\section*{LLM Usage Statement}
The research questions, hypotheses, and conceptual framework were developed by the authors without generative AI assistance. We used generative AI tools, primarily OpenAI Codex, to discuss and refine the methodology and to carry out most of the code implementation. AI tools also assisted with recording and analyzing experimental logs, analyzing and interpreting experimental results, conducting the literature survey, polishing the manuscript and generating visual materials, including background illustrations. The authors reviewed the entire manuscript and the outputs produced by the code, and rigorously tested the implementation against the intended requirements. The authors take responsibility for the final text, claims, results, and research artifacts, including those produced with AI assistance.

\clearpage
\bibliographystyle{assets/plainnat}
\bibliography{paper}

\end{document}